\documentclass[letterpaper]{article} 
\usepackage{aaai25}  
\usepackage{times}  
\usepackage{helvet}  
\usepackage{courier}  
\usepackage[hyphens]{url}  
\usepackage{graphicx} 
\usepackage{natbib}  
\usepackage{caption} 
\usepackage{algorithm}
\usepackage{algorithmic}
\usepackage{multirow}
\usepackage{graphicx}
\usepackage[dvipsnames,table]{xcolor}

\usepackage{float}

\definecolor{White}{rgb}{1.,0.,1.}
\definecolor{first}{rgb}{.8,.0,.0}
\definecolor{second}{rgb}{.0,.6,.0}
\definecolor{third}{rgb}{.0,.0,.8}

\definecolor{ceiling}{RGB}{214,  38, 40}
\definecolor{floor}{RGB}{43, 160, 4}
\definecolor{wall}{RGB}{158, 216, 229}
\definecolor{window}{RGB}{114, 158, 206}
\definecolor{chair}{RGB}{204, 204, 91}
\definecolor{bed}{RGB}{255, 186, 119}
\definecolor{sofa}{RGB}{147, 102, 188}
\definecolor{table}{RGB}{30, 119, 181}
\definecolor{tvs}{RGB}{160, 188, 33}
\definecolor{furniture}{RGB}{255, 127, 12}
\definecolor{objects}{RGB}{196, 175, 214}

\definecolor{car}{rgb}{0.39215686, 0.58823529, 0.96078431}
\definecolor{bicycle}{rgb}{0.39215686, 0.90196078, 0.96078431}
\definecolor{motorcycle}{rgb}{0.11764706, 0.23529412, 0.58823529}
\definecolor{truck}{rgb}{0.31372549, 0.11764706, 0.70588235}
\definecolor{othervehicle}{rgb}{0.39215686, 0.31372549, 0.98039216}
\definecolor{person}{rgb}{1.        , 0.11764706, 0.11764706}
\definecolor{bicyclist}{rgb}{1.        , 0.15686275, 0.78431373}
\definecolor{motorcyclist}{rgb}{0.58823529, 0.11764706, 0.35294118}
\definecolor{road}{rgb}{1.        , 0.        , 1.        }
\definecolor{parking}{rgb}{1.        , 0.58823529, 1.        }
\definecolor{sidewalk}{rgb}{0.29411765, 0.        , 0.29411765}
\definecolor{otherground}{rgb}{0.68627451, 0.        , 0.29411765}
\definecolor{building}{rgb}{1.        , 0.78431373, 0.        }
\definecolor{fence}{rgb}{1.        , 0.47058824, 0.19607843}
\definecolor{vegetation}{rgb}{0.        , 0.68627451, 0.        }
\definecolor{trunk}{rgb}{0.52941176, 0.23529412, 0.        }
\definecolor{terrain}{rgb}{0.58823529, 0.94117647, 0.31372549}
\definecolor{pole}{rgb}{1.        , 0.94117647, 0.58823529}
\definecolor{trafficsign}{rgb}{1.        , 0.        , 0.        }
\definecolor{otherstructure}{rgb}{0.98039215, 0.58823529, 0.}
\definecolor{otherobject}{rgb}{0.19607843, 1.        , 1.        }

\makeatletter
\newcommand{\car@semkitfreq}{3.92}
\newcommand{\bicycle@semkitfreq}{0.03}
\newcommand{\motorcycle@semkitfreq}{0.03}
\newcommand{\truck@semkitfreq}{0.16}
\newcommand{\othervehicle@semkitfreq}{0.20}
\newcommand{\person@semkitfreq}{0.07}
\newcommand{\bicyclist@semkitfreq}{0.07}
\newcommand{\motorcyclist@semkitfreq}{0.05}
\newcommand{\road@semkitfreq}{15.30}
\newcommand{\parking@semkitfreq}{1.12}
\newcommand{\sidewalk@semkitfreq}{11.13}
\newcommand{\otherground@semkitfreq}{0.56}
\newcommand{\building@semkitfreq}{14.1}
\newcommand{\fence@semkitfreq}{3.90}
\newcommand{\vegetation@semkitfreq}{39.3}
\newcommand{\trunk@semkitfreq}{0.51}
\newcommand{\terrain@semkitfreq}{9.17}
\newcommand{\pole@semkitfreq}{0.29}
\newcommand{\trafficsign@semkitfreq}{0.08}
\newcommand{\semkitfreq}[1]{{\csname #1@semkitfreq\endcsname}}

\newcommand{\car@sscbkitfreq}{2.85}
\newcommand{\bicycle@sscbkitfreq}{0.01}
\newcommand{\motorcycle@sscbkitfreq}{0.01}
\newcommand{\truck@sscbkitfreq}{0.16}
\newcommand{\othervehicle@sscbkitfreq}{5.75}
\newcommand{\person@sscbkitfreq}{0.02}
\newcommand{\road@sscbkitfreq}{14.98}
\newcommand{\parking@sscbkitfreq}{2.31}
\newcommand{\sidewalk@sscbkitfreq}{6.43}
\newcommand{\otherground@sscbkitfreq}{2.05}
\newcommand{\building@sscbkitfreq}{15.67}
\newcommand{\fence@sscbkitfreq}{0.96}
\newcommand{\vegetation@sscbkitfreq}{41.99}
\newcommand{\terrain@sscbkitfreq}{7.10}
\newcommand{\pole@sscbkitfreq}{0.22}
\newcommand{\trafficsign@sscbkitfreq}{0.06}
\newcommand{\otherstructure@sscbkitfreq}{4.33}
\newcommand{\otherobject@sscbkitfreq}{0.28}
\newcommand{\sscbkitfreq}[1]{{\csname #1@sscbkitfreq\endcsname}}
\usepackage{amsmath}
\usepackage{amssymb}
\usepackage{booktabs}
\usepackage{adjustbox}
\usepackage[dvipsnames]{xcolor}
\usepackage{newfloat}
\usepackage{listings}
\DeclareCaptionStyle{ruled}{labelfont=normalfont,labelsep=colon,strut=off} 
\floatstyle{ruled}
\newfloat{listing}{tb}{lst}{}
\floatname{listing}{Listing}
\title{LOMA: Language-assisted Semantic Occupancy Network via Triplane Mamba}
\author{
    Yubo Cui,
    Zhiheng Li,
    Jiaqiang Wang,
    Zheng Fang\thanks{Corresponding author.}
}
\affiliations{
    Faculty of Robot Science and Engineering\\
    Northeastern University\\

    \{ybcui21, zhli24, wjq013\}@stumail.neu.edu.cn, fangzheng@mail.neu.edu.cn
}

\begin{document}

\maketitle

\begin{abstract}
Vision-based 3D occupancy prediction has become a popular research task due to its versatility and affordability. 
Nowadays, conventional methods usually project the image-based vision features to 3D space and learn the geometric information through the attention mechanism, enabling the 3D semantic occupancy prediction. 
However, these works usually face two main challenges: 1) Limited geometric information. Due to the lack of geometric information in the image itself, it is challenging to directly predict 3D space information, especially in large-scale outdoor scenes. 2) Local restricted interaction. Due to the quadratic complexity of the attention mechanism, they often use modified local attention to fuse features, resulting in a restricted fusion.
To address these problems, in this paper, we propose a language-assisted 3D semantic occupancy prediction network, named LOMA. 
In the proposed vision-language framework, we first introduce a VL-aware Scene Generator (VSG) module to generate the 3D language feature of the scene.
By leveraging the vision-language model, this module provides implicit geometric knowledge and explicit semantic information from the language.
Furthermore, we present a Tri-plane Fusion Mamba (TFM) block to efficiently fuse the 3D language feature and 3D vision feature. The proposed module not only fuses the two features with global modeling but also avoids too much computation costs.
Experiments on the SemanticKITTI and SSCBench-KITTI360 datasets show that our algorithm achieves new state-of-the-art performances in both geometric and semantic completion tasks. Our code will be open soon.
\end{abstract}

%

\section{Introduction}
In recent years, the 3D scene understanding of autonomous driving has received more and more attention. In order to plan navigation and avoid obstacles more safely, the autonomous driving system needs to perceive the surrounding 3D environment, especially in predicting the occupancy status of the vicinity. However, due to the complexity of the scene, such as occlusion and interference, it is very difficult to predict occupancy in real-world scenes.

In order to address these above challenges, the 3D semantic occupancy prediction task~\cite{behley2019semantickitti}, also denoted as Semantic Scene Completion (SSC), was proposed to simultaneously predict occupancy and semantic information in 3D space. 
Meanwhile, the visual image is often employed to predict 3D semantic occupancy due to its affordability and capacity to provide detailed visual information.

\begin{figure}[t]
\centering
\includegraphics[width=0.9\linewidth]{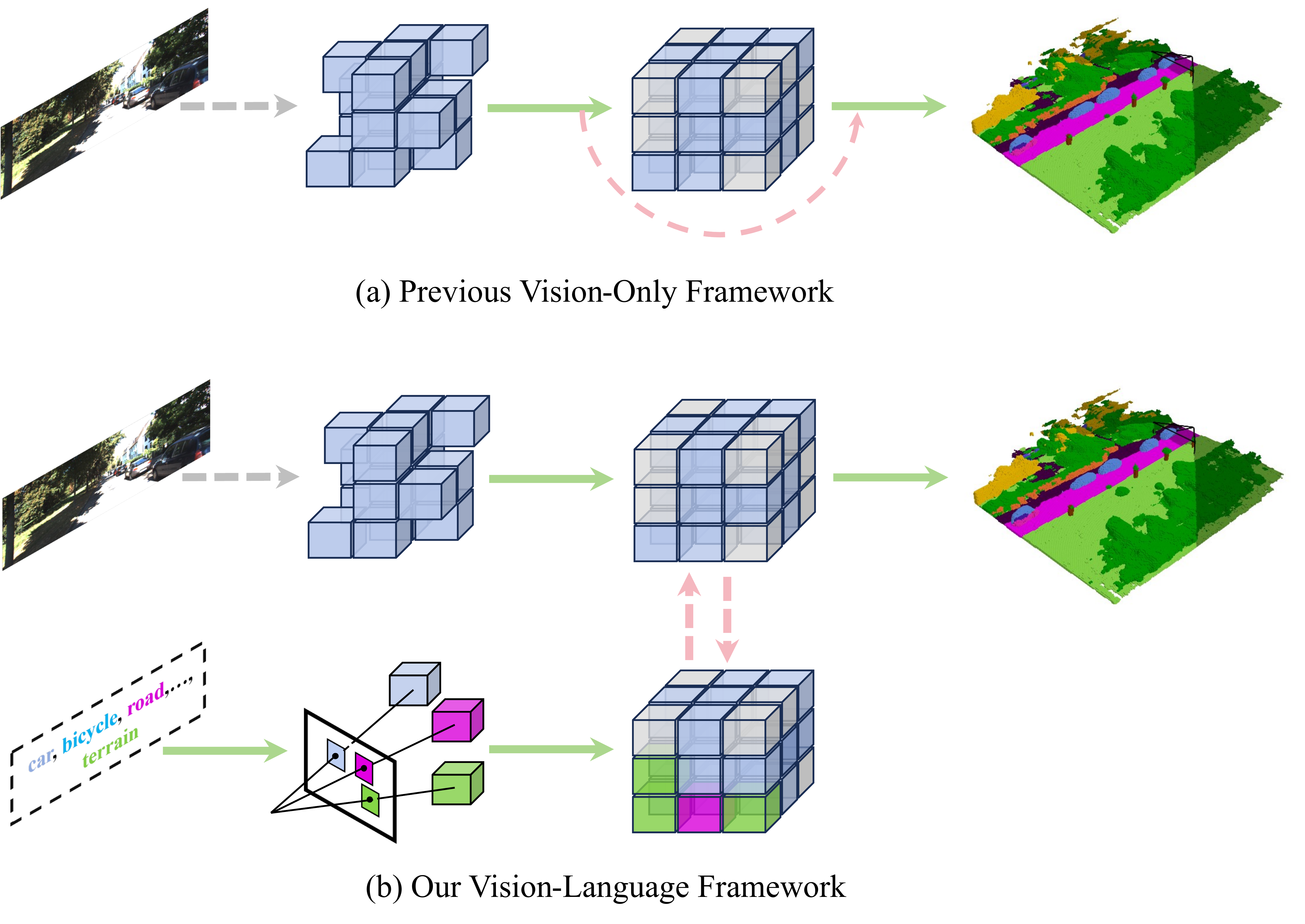}
\caption{(a) Previous vision-only framework. (b) Our proposed vision-language framework. Compared to (a), our method introduces the explicit prior from language to enhance the 3D scene understanding.\textbf{\textcolor{gray}{$\dashrightarrow$}} and \textcolor{CarnationPink}{$\dashrightarrow$} represents \textit{2D-to-3D} and \textit{3D-to-3D} feature propagation respectively.}
\label{fig:intro}
\end{figure}

MonoScene~\cite{cao2022monoscene} laid the foundation for vision-based 3D semantic occupancy prediction. By translating multi-scale 2D image features into 3D spaces based on the camera intrinsic matrix, they developed a framework for predicting semantic occupancy in 3D scenes based on 2D images.
Based on the ``first \textit{2D-to-3D} then \textit{3D-to-3D}'' feature propagation pipeline, further advancements have been made to enhance prediction accuracy.
For example, TPVFormer~\cite{huang2023tri} projects each 3D point into three orthogonal planes and utilizes the attention mechanism to obtain 3D features from 2D images. 
OccFormer~\cite{zhang2023occformer} breaks down the intensive 3D processing into local and global transformers, introducing the dual-path transformer block to better capture intricate details and overall scene structures.

Despite notable advancements, most existing methods concentrate on improving the overall performance by integrating image information into dense 3D scenes, as shown in Figure.~\ref{fig:intro}(a). 
However, this vision-only framework faces two challenges: 
1) \textit{Limited geometric information.} While images contain rich visual texture information, they have limited geometric information. Although pre-trained depth prediction networks can provide depth estimation, it is still difficult for the image to predict the 3D semantics and occupancy simultaneously.
2) \textit{Local restricted interaction.} 
Due to the high computational complexity of 3D features, previous transformer-based methods usually perform a limited 3D-to-3D feature propagation by some modified attention, \textit{i.e.} deformable attention~\cite{DeformableDETR}.
Although these methods have limited computational costs, they also lose the global feature integration.

To this end, in this paper, we propose a \textbf{L}anguage-assisted 3D semantic \textbf{O}ccupancy network via triplane \textbf{MA}mba, called LOMA. 
Specifically, we first propose the VL-aware Scene Generator (VSG) module to provide the 3D voxel-wise language feature. 
Compared to image, language contains explicit semantics and implicit geometric information. For instance, the word ``cars" evokes an immediate virtual image of a vehicle occupying roughly $4m \times 2m \times 2m$ in 3D space, even if we are only observing a portion of the actual car.
Therefore, we can utilize the prior information from language to improve the geometric prediction of 3D scenes and assist in semantic occupancy prediction.
Moreover, drawing inspiration from the recent advancements in State Space Models (SSMs), we then present the Tri-plane Fusion Mamba (TFM) module to perform 3D-to-3D feature propagation. This module not only performs feature propagation from non-empty voxels to empty voxels, but also propagates different modalities between vision and language.
By projecting 3D scene features onto three mutually orthogonal planes, our proposed module conducts global feature interaction on each plane and updates vision and language features simultaneously.
Furthermore, we extend this approach to a multi-scale manner for more comprehensive feature interaction.
In comparison to the attention mechanism, our SSM-based approach allows for global feature interaction while reducing computational burden.

Our contributions can be summarized as follows:
\begin{itemize}
    \item We propose a novel vision-language framework, which efficiently utilizes the prior from language to assist in vision-based 3D semantic occupancy prediction.
    \item We present VL-aware Scene Generator (VSG) module and Tri-plane Fusion Mamba (TFM) module to introduce the voxel-wise 3D language feature and perform 3D-to-3D feature propagation, respectively.
    \item We compare the proposed LOMA on SemanticKITTI and SSCBench-KITTI360 datasets and show our method outperforms previous state-of-the-art approaches.
\end{itemize}

\section{Related Works}
\subsection{3D Semantic Occupancy Prediction}
SSCNet~\cite{SSCNet} first proposes the 3D semantic occupancy prediction task, also known as the semantic scene completion (SSC) task. Taking the depth map as input, they jointly predict the volumetric occupancy and semantic labels for full 3D scenes simultaneously.
MonoScene~\cite{cao2022monoscene} first introduces the 2D image into the 3D semantic occupancy prediction task. They lift 2D visual features to 3D space and capture long-range semantic context, resulting in a good performance.
Nowadays, several studies~\cite{wei2023surroundocc, zheng2024monoocc} utilize attention to perform feature propagation.
Voxformer~\cite{li2023voxformer} adopts a two-stage approach to perform 2D-to-3D and 3D-to-3D feature propagation, reducing the huge computational burden caused by dense 3D features.
To alleviate the limitations caused by geometric ambiguity, Symphonies~\cite{jiang2024symphonize} proposes an instance-centric method to predict through instance-scene propagation.
HASSC~\cite{hassc} follows the principle that not all voxels are equal and introduces hardness-aware semantic scene completion. 
Compared to these vision-only approaches, we introduce language into this task and propose a novel vision-language framework to predict the 3D semantic occupancy.

\begin{figure*}[t]
\centering
\includegraphics[width=0.9\linewidth]{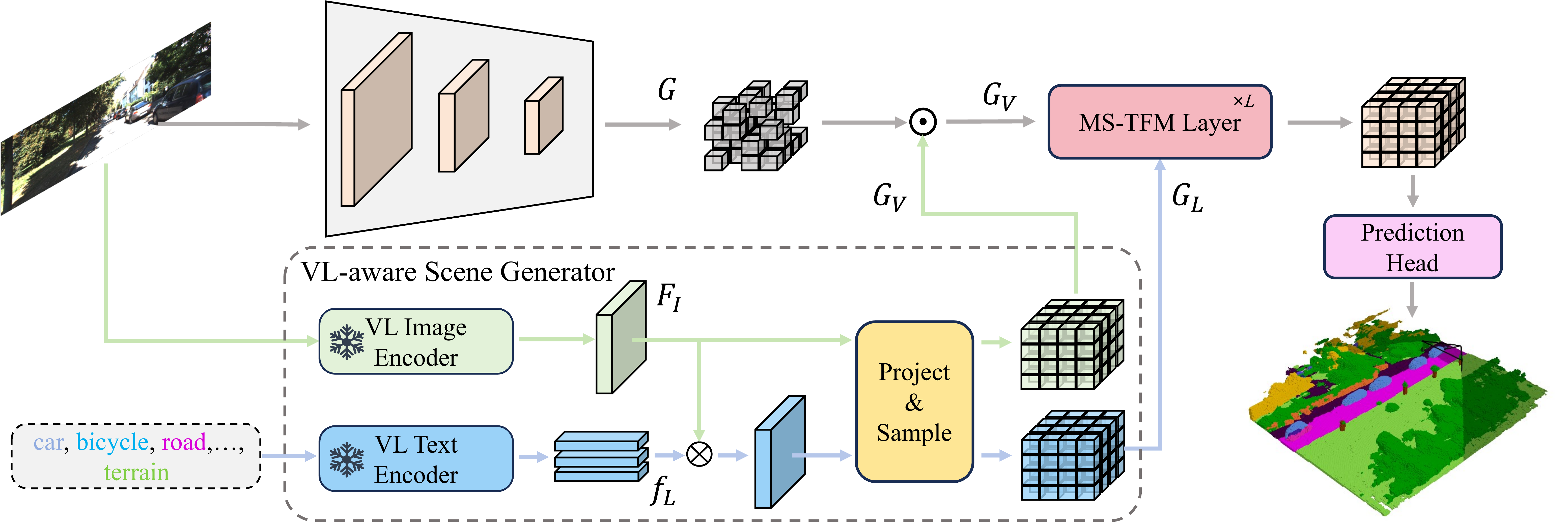}
\caption{Architecture of the proposed LOMA. We input the image and categories text as inputs. The image encoder extracts multi-scale image features from the image and performs 2D-to-3D feature propagation through deformable attention. Meanwhile, the VL-aware scene generator utilizes VLM to generate the scene-level 3D features. We further propose the Multi-scale Triplane Fusion Mamba (MS-TFM) layer to fuse the 3D scene-level vision and language features. Finally, the fused vision feature is used to predict the semantic occupancy. For clarity, the pre-trained depth network is omitted.}
\label{fig:arch}
\end{figure*}

\subsection{Vision-Language Models}
Recently, Vision-Language Models (VLMs) have made significant progress due to their broad applications and multi-modal capabilities. By constructing image-text pairs, VLMs use an image encoder and text encoder to establish the vision-language correlation. 
CLIP~\cite{clip} is the notable milestone in this field, which employs contrastive pre-training between image and language encoders, exhibiting impressive zero-shot classification performance. However, CLIP requires carefully crafted prompts in the text encoder, posing a challenge in the formulation.
To tackle this issue, CoOp~\cite{coop} and CoCoOp~\cite{cocoop} train dynamic soft-prompt during training and condition on image input.
Moreover, some studies extend vision-language models for pixel-wise semantic segmentation, aiming at open-vocabulary segmentation.
On one hand, LSeg~\cite{lseg} utilized CLIP to train pixel-level visual embeddings that align with the text embeddings of CLIP.
On the other hand, OpenSeg~\cite{openseg} suggests detecting specific local areas in images and establishing correlations with text embeddings through class-agnostic region proposals. 
To enhance the identification of these areas, MaskCLIP~\cite{maskclip} leverages the self-attention map from CLIP to improve the precision of region proposals.
In our study, we leverage VLMs to provide language features for 3D semantic occupancy prediction.

\subsection{State Space Models}
The State Space Model (SSM) was used to describe dynamic systems in modern control theory. Some previous works~\cite{ssm1, ssm2, ssm3} have introduced it into the field of deep learning, as an architectural paradigm for sequence-to-sequence transformations. 
Recent works have made significant progress, making deep SSM a powerful competitor against CNN and Transformers.
In particular, S4~\cite{s4} presents a practical Normal Plus Low-Rank (NPLR) method to speed up matrix inversion, making the convolution kernel computation more efficient.
S5~\cite{s5} presents the parallel scan and the MIMO SSM, enabling the effective utilization of the state space model.
More recently, Mamba~\cite{mamba} presents input-dependent SSMs and develops a versatile framework that competes well with finely tuned Transformers. 
Inspired by the success of Mamba, some studies~\cite{vim, vmamba, localmamba, multiscalemamba} have begun to expand the mamba into vision tasks. For example, VMamba~\cite{vmamba} proposes cross-selective scanning mechanisms to compensate for the difference between 1D sequences and 2D images. Meanwhile, Vim~\cite{vim} introduces a bidirectional state space modeling or capturing data-dependent global visual context. In this paper, we investigate the utilization of SSMs to propagate 3D features efficiently.

\section{Methodology}
\subsection{Overview}
The architecture of the proposed LOMA is shown in Figure.~\ref{fig:arch}. Similar to the previous methods~\cite{li2023voxformer, jiang2024symphonize}, we first feed the image into an image encoder to obtain multi-scale features. Meanwhile, based on the depth map through a pre-trained depth model, we use the deformable attention~\cite{DeformableDETR} to fuse the 2D multi-scale features with a pre-defined learnable 3D feature $G \in \mathbb{R}^{HWL\times D}$, performing 2D-to-3D feature propagation.
Furthermore, different from the previous vision-only framework, our proposed vision-language framework also takes the category text as input. We input the image and category text into the \textit{VL-aware Scene Generator} (VSG) module, which leverages the VLM to incorporate language information. 
The proposed module generates 3D VL-vision features $G_V$ and 3D VL-language features $G_L$ for images and corresponding language. Since they both come from the visual image, we directly concatenate $G_V$ and $G$ and use 3D convolution layers to fuse them, achieving the final 3D vision features $G_V$.
Finally, we integrate the 3D language feature $G_L$ and the 3D vision feature $G_V$ through the proposed \textit{Multi-scale Tri-plane Fusion Mamba} (MS-TFM) layers, and use the fused 3D vision features for semantic occupancy prediction of the entire 3D scene. 

\begin{figure*}[t]
\centering
\includegraphics[width=0.9\linewidth]{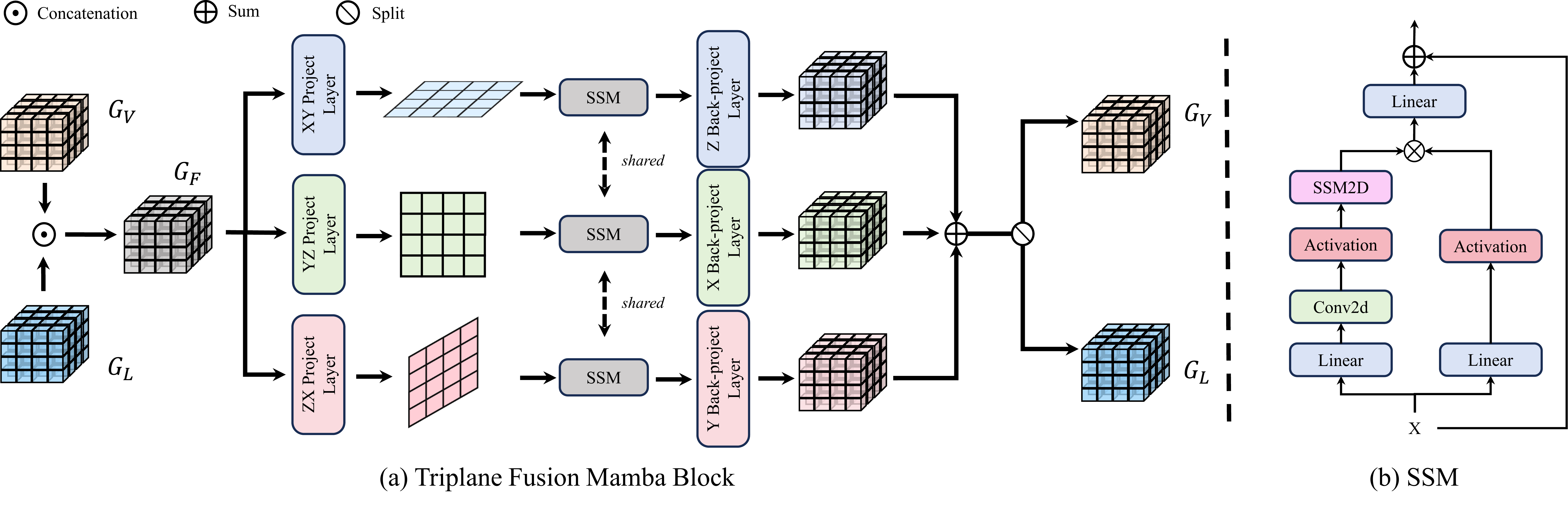}
\caption{(a) Architecture of the proposed TFM module. We concatenate the two different modality features along the feature channel and use three Linear layers to project it to three 2D plane features, respectively. Then, a shared SSM block is used to perform global interaction. Subsequently, we use three Linear layers to project the 2D features back to 3D features and sum them up. Finally, the updated vision and language features are split along the feature channel.
(b) The detail of the SSM block.}
\label{fig:tmf}
\end{figure*}

\subsection{VL Scene Generator}
To introduce language priors in 3D semantic occupancy prediction, it is crucial to establish a bridge between images and language. Benefiting from the large-scale image-text pairs, the vision-language models (VLMs) could generate robust language priors for the corresponding image. 
In this module, we use existing pre-trained VLM to integrate language priors. Specifically, given the input image $I \in \mathbb{R}^{H_I\times W_I\times 3}$ and categories text $T \in \mathbb{R}^{N}$, where $N$ is the number of categories and $H_I\times W_I$ represents the resolution of the image, we extract the image feature $F_I\in \mathbb{R}^{H_I\times W_I\times C}$ and language feature $f_L \in \mathbb{R}^{N\times C}$ from the VL image encoder and VL text encoder. 
Then, to build the pixel-level language feature, we first establish the pixel-level category as follows:
\begin{equation}
    M=\textbf{argmax}\left(\text{softmax}\left(\frac{F_I \cdot f_L}{t}\right)\right)
\end{equation}
where $\cdot$ represents the inner product and $t$ is a pre-defined temperature parameter. Thus the output $M\in\mathbb{R}^{H_I\times W_I\times 1}$ represents the category index of each pixel.
Furthermore, we use 2D Conv layer and Linear layer to align the feature channel of $F_I$ and $f_L$ to the the above 3D feature $G$. 

By utilizing the pre-trained VLM, we obtain the 2D image label $M$, 2D image feature $F_I$ and 1D language feature $f_L$. 
However, we expect to make predictions for 3D scenes, therefore we need to convert them to 3D space. Here, we adopt a simple yet effective projection and sampling approach.
Since the defined grids have fixed 3D coordinates, we can convert the voxel grid coordinate to image pixel coordinates based on the known camera parameters, and query the 3D feature by nearest sampling the 2D feature based on the projected coordinates. For a given voxel grid with 3D coordinate $x_w$, its 3D VL-vision feature $g_v \in \mathbb{R}^{D}$ and VL-language feature $g_l \in \mathbb{R}^{D}$ could be formulated as follows:
\begin{align}
    g_v&=s(F_I, \pi(x_w))\\
    g_l&=f_L(s(M, \pi(x_w)))
\end{align}
where $\pi, s$ represent world-to-image transformation and sampling respectively. With this approach, we can transform 2D features into 3D space without introducing too much computational complexity. 
Finally, through the proposed VSG, we generate the 3D scene-level features, providing the language information for the subsequent 3D-to-3D feature propagation.

\subsection{Multi-scale Tri-plane Fusion Mamba Layer}
Although previous works do not involve the fusion of 3D features from different modalities, they typically use the self-attention mechanisms to the 3D vision features for feature diffusion, performing 3D-to-3D feature propagation. While attention performs global feature fusion, it also brings significant computational complexity, \textit{i.e.} $\mathcal{O}(n^2)$, especially for 3D features. Some works alleviate this problem by reducing the number of keys in attention computation, but this approach also leads to local perception.
In this paper, we introduce Mamba for 3D-to-3D feature propagation between different modalities. 
Compared to attention mechanisms, Mamba incurs a lower computational cost, \textit{i.e.} $\mathcal{O}(n)$, while also offering global perception capability, presenting a novel approach for 3D feature propagation.

\subsubsection{Preliminaries.}
The state space sequence (SSM) model is a continuous system that maps 1D inputs $x(t)$ to outputs $y(t)$ through hidden states, which can be represented as follows:
\begin{equation}
\begin{aligned}
h'(t) &= \mathbf{A}h(t)+\mathbf{B}x(t)\\
y(t) &= \mathbf{C}h(t)+\mathbf{D}x(t)
\end{aligned}
\end{equation}
where $\mathbf{A} \in \mathbb{R}^{N\times N}, \mathbf{B} \in \mathbb{R}^{N\times 1}, \mathbf{C} \in \mathbb{R}^{1\times N}$ are learnable parameters. $\mathbf{D}\in \mathbb{R}^1$ denotes a residual connection.
To integrate it into deep learning, it is necessary to discretize the above continuous systems. By assuming a timescale parameter $\Delta$, the discrete parameters can be represented as follows:
\begin{equation}
\begin{aligned}
    \overline{\mathbf{A}} &= \exp (\Delta \mathbf{A})\\
    \overline{\mathbf{B}} &= (\Delta \mathbf{A})^{-1}(\exp (\Delta \mathbf{A})-\mathbf{I})\cdot \Delta \mathbf{B}
\end{aligned}
\end{equation}
Thus, the overall system could be discretized as follows:
\begin{equation}
\begin{aligned}
    & h_k = \overline{\mathbf{A}}h_{k-1}+\overline{\mathbf{B}}x_k\\
    & y_k = \overline{\mathbf{C}}h_k+\overline{\mathbf{D}}x_k
\end{aligned}
\end{equation}
Finally, a global convolution is used for parallel processing:
\begin{align}
    \overline{\mathbf{K}} &= (\mathbf{C}\overline{\mathbf{B}}, \mathbf{C}\overline{\mathbf{A}\mathbf{B}},...,\mathbf{C}\overline{\mathbf{A}}^{M-1}\overline{\mathbf{B}})\\
    \mathbf{y} &= \mathbf{x} \cdot \overline{\mathbf{K}}
\end{align}
where $M$ is the length of the input $\mathbf{x}$. $\overline{\mathbf{K}}$ is the structured convolution kernel.
\subsubsection{Triplane Fusion Mamba Block.}
Given 3D vision feature $G_V \in \mathbb{R}^{H\times W\times L \times D}$ and 3D language feature $G_L \in \mathbb{R}^{H\times W\times L \times D}$. We aim to integrate the prior information from language into the vision features. 

We first concatenate the two 3D features along the feature channel to get the fusion feature $G_F \in \mathbb{R}^{H\times W\times L \times 2C}$. 
However, despite reducing the computational complexity from $O((HWL)^2)$ to $O(HWL)$ with Mamba, there is still a significant computational burden due to the large size of 3D scene features.
Therefore, we propose the Tri-plane Fusion Mamba (TFM) block. 
By integrating the three axes into the feature channel and projecting them onto three 2D planes, we can significantly reduce the computational load without losing information.
Specifically, as shown in Figure.~\ref{fig:tmf}, we project the 3D fusion features $G_F$ through three Linear layers onto the XY, YZ and ZX planes respectively, resulting in three 2D features. We then utilize a shared SSM block to process these 2D features, performing vision-language feature interaction as follows:
\begin{align}
    F^{XY} &= \text{SSM}(\text{Linear}_{XY}(G_F)), F^{XY}\in \mathbb{R}^{H\times W\times D}\\
    F^{YZ} &= \text{SSM}(\text{Linear}_{YZ}(G_F)), F^{YZ}\in \mathbb{R}^{W\times L\times D}\\
    F^{ZX} &= \text{SSM}(\text{Linear}_{ZX}(G_F)), F^{ZX}\in \mathbb{R}^{L\times H\times D}
\end{align}
After that, we further use Linear layer on each 2D plane feature to restore it to a 3D feature, and sum them up to get the final fused feature. Thus, it could be represented as follows:
\begin{align}
    & G^{XY}= \text{Linear}_{z}(F^{XY}), G^{XY}\in \mathbb{R}^{H\times W\times LD}\\
    & G^{YZ}= \text{Linear}_{x}(F^{YZ}), G^{YZ}\in \mathbb{R}^{W\times L\times HD}\\
    & G^{ZX}= \text{Linear}_{y}(F^{ZX}), G^{ZX}\in \mathbb{R}^{L\times H\times WD}\\
    & G_F= G^{XY} + G^{YZ} + G^{ZX}
\end{align}
For clarity, we omit the shape alignment of features in the equations. Since we concatenated the vision and language features along the feature channel before, we then split the two features back, leading to updated 3D features.

Through the proposed TFM block, we not only perform 3D-to-3D feature propagation between vision features and language features in a global manner but also perform feature propagation from non-empty voxel to empty voxel, leading to a more comprehensive feature interaction.
Moreover, the introduction of SSM and tri-plane further reduces the computation complexity, alleviating the issue of high computational costs from 3D features.

\begin{figure}[t]
\centering
\includegraphics[width=0.9\linewidth]{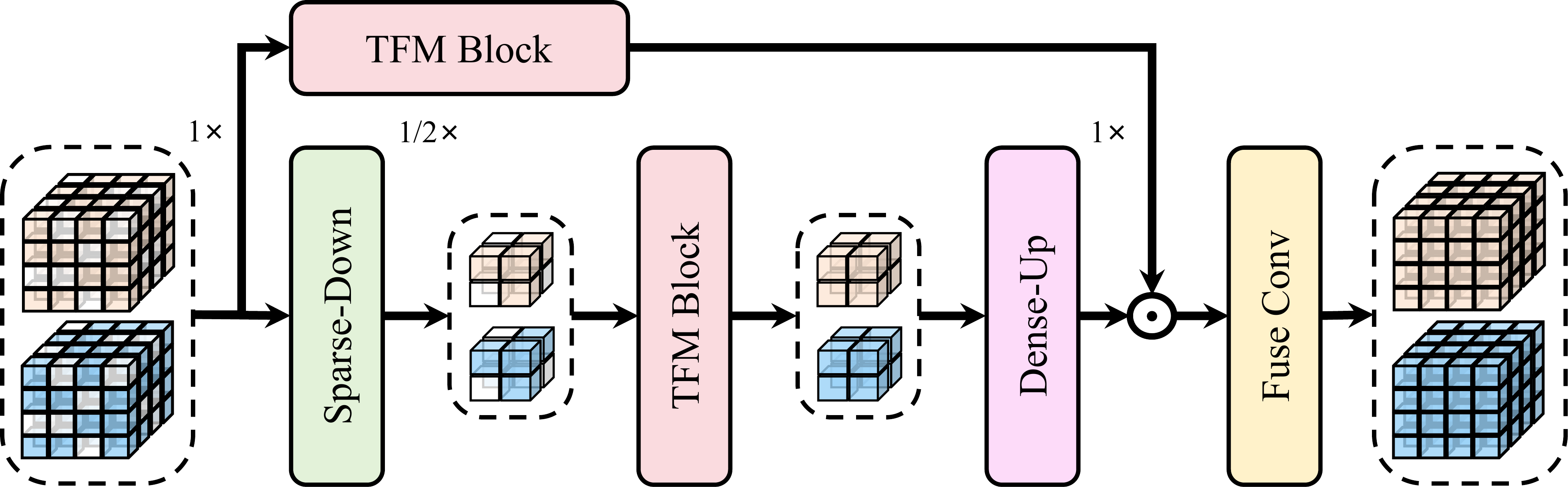}
\caption{Illustration of the proposed MS-TFM layer.}
\label{fig:ms_tmf}
\end{figure}
\subsubsection{Extending to Multi-scale.}
In order to make better use of high-level information, we extend TFM to a multi-scale approach. 
For this, we need to extract 3D features at different scales.
However, our 3D vision and language features are both derived from image features through 2D-to-3D feature propagation, with only a small portion of the 3D voxel grid involved. Therefore, the 3D features are sparse and the majority of voxels are kept empty.
Based on this observation, we employ 3D sparse convolution in the down-sampling. 

Specifically, as shown in Figure.~\ref{fig:ms_tmf}, similar to the processing of point clouds, we use a sparse convolution layer and two submanifold convolution layers to extract high-level features. The sparse convolution layer processes feature for all voxels, whether empty or not, while the submanifold convolution layer only processes feature at non-empty voxels.
Then, for 3D vision and language features at each scale, we use the proposed TFM block to perform 3D-to-3D feature propagation.
Since TFM also propagates features from non-empty voxels to empty voxels, we get a dense representation after the module.
Therefore, we use the regular 3D deconvolution layer for up-sampling and merge them with low-level 3D features, outputting the fused vision and language features by a regular 3D Conv layer. 


\section{Experiments}
\subsection{Experimental Setup}
\subsubsection{Dataset.}
Following previous works~\cite{jiang2024symphonize}, we evaluate the proposed LOMA on SemanticKITTI~\cite{behley2019semantickitti} and SSCBench-KITTI360~\cite{li2023sscbench} datasets. SemanticKITTI comprises 22 driving sequences, with an official split of 10, 1, and 11 sequences for training, validation, and testing respectively. The input RGB images are with sizes of $1226\times 370$, and the annotation label has 20 semantic classes. The output scene covers an area of $51.2m\times 51.2m\times 64m$ and is voxelized into a grid with a shape of $256\times 256\times 32$ using voxels of size 0.2m. SSCBench-KITTI360 includes 7 training sequences, 1 validation sequence and 1 testing sequence. Its input RGB images are with sizes of $1408\times 376$, and the annotation label has 19 semantic classes. 
SSCBench-KITTI360 also has the voxel size of $256\times 256\times 32$.

\subsubsection{Evaluation Metrics.}
Following previous works~\cite{li2023voxformer}, we utilize the mean IoU (mIoU) metrics to assess the semantic prediction accuracy and employ the intersection over union (IoU) to measure the prediction accuracy.

\subsection{Implementation Details}
In our implementation, we use Resnet-50~\cite{resnet} to extract multi-scale visual features and use the LSeg~\cite{lseg} model as our VLM to extract vision-language features from image and text. Following previous works, we also adopt the pre-trained MobileStereoNet~\cite{mobilestereonet} to estimate the depth map. 
We utilize the AdamW optimizer with an initial learning rate of $2\times10^{-4}$ and a weight decay of $10^{-4}$. We train our LOMA for 30 epochs on 4 NVIDIA 3090 GPUs, with a batch size of 4, and employ random horizontal flip augmentations.
\begin{table*}[ht]
\centering
\newcommand{\clsname}[2]{
    \rotatebox{90}{
        \hspace{-6pt}
        \textcolor{#2}{$\blacksquare$}
        \hspace{-6pt}
        \renewcommand\arraystretch{0.6}
        \begin{tabular}{l}
            #1                                      \\
            \hspace{-4pt} ~\tiny(\semkitfreq{#2}\%) \\
        \end{tabular}
    }}
\renewcommand{\tabcolsep}{2.0pt}
\renewcommand\arraystretch{0.8}
\scalebox{0.8}
{
\begin{tabular}{l|r|r>{\columncolor{gray!20}}r|rrrrrrrrrrrrrrrrrrrr}
    \toprule[.05cm]
    Method                               &
    Source                               &
    \multicolumn{1}{c}{IoU}              &
    mIoU                                 &
    \clsname{road}{road}                 &
    \clsname{sidewalk}{sidewalk}         &
    \clsname{parking}{parking}           &
    \clsname{other-grnd.}{otherground}   &
    \clsname{building}{building}         &
    \clsname{car}{car}                   &
    \clsname{truck}{truck}               &
    \clsname{bicycle}{bicycle}           &
    \clsname{motorcycle}{motorcycle}     &
    \clsname{other-veh.}{othervehicle}   &
    \clsname{vegetation}{vegetation}     &
    \clsname{trunk}{trunk}               &
    \clsname{terrain}{terrain}           &
    \clsname{person}{person}             &
    \clsname{bicyclist}{bicyclist}       &
    \clsname{motorcyclist}{motorcyclist} &
    \clsname{fence}{fence}               &
    \clsname{pole}{pole}                 &
    \clsname{traf.-sign}{trafficsign}
    \\
    \midrule
    LMSCNet$^\dagger$  &3DV'20   & 31.38          & 7.07           & 46.70          & 19.50          & 13.50          & 3.10           & 10.30          & 14.30          & 0.30          & 0.00          & 0.00          & 0.00          & 10.80          & 0.00           & 10.40          & 0.00          & 0.00          & 0.00          & 5.40           & 0.00          & 0.00          \\
    AICNet$^\dagger$  & CVPR'20    & 23.93          & 7.09           & 39.30          & 18.30          & 19.80          & 1.60           & 9.60           & 15.30          & 0.70          & 0.00          & 0.00          & 0.00          & 9.60           & 1.90           & 13.50          & 0.00          & 0.00          & 0.00          & 5.00           & 0.10          & 0.00          \\
    JS3C-Net$^\dagger$ &AAAI'21   & 34.00          & 8.97           & 47.30          & 21.70          & 19.90          & 2.80           & 12.70          & 20.10          & 0.80          & 0.00          & 0.00          & 4.10          & 14.20          & 3.10           & 12.40          & 0.00          & 0.20          & 0.20          & 8.70           & 1.90          & 0.30          \\
    MonoScene$^\ast$&CVPR'22    & 34.16          & 11.08          & 54.70          & 27.10          & 24.80          & 5.70           & 14.40          & 18.80          & 3.30          & 0.50          & 0.70          & 4.40          & 14.90          & 2.40           & 19.50          & 1.00          & 1.40          & 0.40          & 11.10          & 3.30          & 2.10          \\
    TPVFormer    &CVPR'23     & 34.25          & 11.26          & 55.10          & 27.20          & 27.40          & 6.50           & 14.80          & 19.20          & 3.70 & 1.00          & 0.50          & 2.30          & 13.90          & 2.60           & 20.40          & 1.10          & 2.40          & 0.30          & 11.00          & 2.90          & 1.50          \\
    VoxFormer   &CVPR'23    & 42.95 & 12.20          & 53.90          & 25.30          & 21.10          & 5.60           & 19.80          & 20.80          & 3.50          & 1.00          & 0.70          & 3.70          & 22.40          & 7.50           & 21.30          & 1.40          & \underline{2.60} & 0.20          & 11.10          & 5.10          & 4.90          \\
    OccFormer  &ICCV'23  & 34.53          & 12.32          & 55.90          & \underline{30.30} & \underline{31.50} & 6.50    & 15.70          & 21.60     & 1.20   & 1.50   & \underline{1.70}    & 3.20          & 16.80    & 3.90    & 21.30    & \underline{2.20}   & 1.10   & 0.20          & 11.90    & 3.80   & 3.70  \\
    SurroundOcc&ICCV'23&34.72 &11.86 &56.90 &28.30 &30.20 &6.80 &15.20 &20.60 &1.40 &1.60 &1.20 &4.40 &14.90 &3.40 &19.30 &1.40 &2.00 &0.10 &11.30 &3.90 &2.40 \\
    MonoOcc &ICRA'24 &- & 13.80 &55.20 &27.80 &25.10 &\underline{9.70} &21.40 &23.20 &\textbf{5.20} &\underline{2.20} &1.50 &5.40 &24.00 &8.70 &23.00 &1.70 &2.00 &0.20 &13.40 &5.80 &\underline{6.40} \\
    Symphonies & CVPR'24    & 42.19          & \underline{15.04} & \textbf{58.40} & 29.30  & 26.90  & \textbf{11.70} &\underline{24.70} & \underline{23.60} & 3.20          & \textbf{3.60} & \textbf{2.60} & \underline{5.60} & \underline{24.20} & \textbf{10.00} &23.10 & \textbf{3.20} & 1.90          & \textbf{2.00} &\underline{16.10} &\textbf{7.70} & \textbf{8.00} \\ 
    HASSC &CVPR'24&\textbf{43.40} &13.34 &54.60 &27.70 &23.80 &6.20 &21.10 &22.80 &\underline{4.70} &1.60 &1.00 &3.90 &23.80 &8.50 &\underline{23.30} &1.60 &\textbf{4.00} &0.30 &13.10 &5.80 &5.50\\ \hline
    Ours &  & \underline{43.01}         & \textbf{15.10} & \underline{57.98} & \textbf{31.80}   & \textbf{32.16}   & 9.47 &\textbf{25.28} & \textbf{24.88} & 4.08  & 1.74 & 1.68 & \textbf{6.36} & \textbf{25.63} &\underline{8.71} & \textbf{24.72} & 1.41&1.74  &\underline{0.64} & \textbf{16.84} &\underline{6.53} &6.08 \\
    \bottomrule[.05cm]
\end{tabular}
}
\caption{\textbf{Quantitative results on SemanticKITTI \texttt{test}.} $^\dagger$ denotes the results provided by MonoScene. $^\ast$ represents the reproduced results in TPVFormer. The best and second results are in \textbf{bold} and \underline{underlined}, respectively.}
\label{tab:sem_kitti_test}
\end{table*}

\begin{table*}[ht]
    \centering
    \newcommand{\clsname}[2]{
        \rotatebox{90}{
            \hspace{-6pt}
            \textcolor{#2}{$\blacksquare$}
            \hspace{-6pt}
            \renewcommand\arraystretch{0.6}
            \begin{tabular}{l}
                #1                                       \\
                \hspace{-4pt} ~\tiny(\sscbkitfreq{#2}\%) \\
            \end{tabular}
        }}
    \newcommand{\empa}[1]{\textbf{#1}}
    \newcommand{\empb}[1]{\underline{#1}}
    \renewcommand{\tabcolsep}{1.8pt}
    \renewcommand\arraystretch{1.0}
    \scalebox{0.8}
    {
    \begin{tabular}{l|rrr>{\columncolor{gray!20}}r|rrrrrrrrrrrrrrrrrr}
        \toprule[.05cm]
        \multicolumn{1}{c|}{Method}                                 &
        \multicolumn{1}{c}{IoU}                                     &
        \multicolumn{1}{c}{Prec.}                                   &
        \multicolumn{1}{c}{Rec.}                                    &
        mIoU                                                        &
        \multicolumn{1}{c}{\clsname{car}{car}}                      &
        \multicolumn{1}{c}{\clsname{bicycle}{bicycle}}              &
        \multicolumn{1}{c}{\clsname{motorcycle}{motorcycle}}        &
        \multicolumn{1}{c}{\clsname{truck}{truck}}                  &
        \multicolumn{1}{c}{\clsname{other-veh.}{othervehicle}}      &
        \multicolumn{1}{c}{\clsname{person}{person}}                &
        \multicolumn{1}{c}{\clsname{road}{road}}                    &
        \multicolumn{1}{c}{\clsname{parking}{parking}}              &
        \multicolumn{1}{c}{\clsname{sidewalk}{sidewalk}}            &
        \multicolumn{1}{c}{\clsname{other-grnd.}{otherground}}      &
        \multicolumn{1}{c}{\clsname{building}{building}}            &
        \multicolumn{1}{c}{\clsname{fence}{fence}}                  &
        \multicolumn{1}{c}{\clsname{vegetation}{vegetation}}        &
        \multicolumn{1}{c}{\clsname{terrain}{terrain}}              &
        \multicolumn{1}{c}{\clsname{pole}{pole}}                    &
        \multicolumn{1}{c}{\clsname{traf.-sign}{trafficsign}}       &
        \multicolumn{1}{c}{\clsname{other-struct.}{otherstructure}} &
        \multicolumn{1}{c}{\clsname{other-obj.}{otherobject}}
        \\
        \hline
        MonoScene    & 37.87        & 56.73        & 53.26        & 12.31        & 19.34        & 0.43        & 0.58        & 8.02         & 2.03         & 0.86        & 48.35        & 11.38        & 28.13        & 3.32        & 32.89        & 3.53        & 26.15        & 16.75        & 6.92         & 5.67        & 4.20         & 3.09         \\
        TPVFormer          & 40.22        & 59.32        & \empb{55.54} & 13.64        & 21.56        & 1.09        & 1.37        & 8.06         & 2.57         & 2.38        & 52.99        & 11.99        & 31.07        & 3.78        & 34.83        & 4.80        & 30.08        & 17.52        & 7.46         & 5.86        & 5.48         & 2.70         \\
        VoxFormer             & 38.76        & 58.52        & 53.44        & 11.91        & 17.84        & 1.16        & 0.89        & 4.56         & 2.06         & 1.63        & 47.01        & 9.67         & 27.21        & 2.89        & 31.18        & 4.97        & 28.99        & 14.69        & 6.51         & 6.92        & 3.79         & 2.43         \\
        OccFormer           & 40.27        & 59.70        & 55.31        & 13.81        & 22.58        & 0.66        & 0.26        & 9.89         & 3.82         & 2.77        & 54.30        & 13.44        & 31.53        & 3.55        & \empb{36.42} & 4.80        & 31.00        & \empb{19.51} & 7.77         & 8.51        & 6.95         & 4.60         \\
        Symphonies                                                & \empb{44.12} & \textbf{69.24} & 54.88       & \empa{18.58} & \textbf{30.02} & \underline{1.85} & \empa{5.90} & \empa{25.07} & \empa{12.06} & \empa{8.20} & \empb{54.94} & \empb{13.83} & \empb{32.76} & \empa{6.93} & 35.11        & \empa{8.58} & \textbf{38.33} & 11.52        & \underline{14.01} & \underline{9.57} & \empa{14.44} & \empa{11.28} \\  \hline
        Ours                                                       & \textbf{46.35} & \underline{64.55} & \textbf{62.17}       & \underline{18.28} & \underline{27.59} & \empa{2.57} & \underline{3.57} & \underline{11.49}& \underline{7.47} & \underline{5.53} & \textbf{58.60} & \textbf{15.76} & \textbf{37.52} & \underline{4.81} & \textbf{41.20}    & \underline{8.42} & \underline{37.72} & \textbf{20.27}        & \empa{14.62} & \empa{16.40} & \underline{8.97} & \underline{6.51} \\
        \bottomrule[.05cm]
        \end{tabular}
    }
    \caption{\textbf{Quantitative results on SSCBench-KITTI360 \texttt{test}.} The results for counterparts are provided in \cite{li2023sscbench}. The best and the second results are in \empa{bold} and \empb{underlined}, respectively.}
    \label{tab:kitti_360_test}
\end{table*}

\begin{table}[t]
\centering
\renewcommand{\tabcolsep}{10pt}
\renewcommand\arraystretch{0.6}
\begin{tabular}{ccc|cc}
\toprule[.05cm]
VSG.      & TFM. & MS.   &  IoU   & mIoU\\
\midrule
&            &            & 40.42     & 13.36\\
& \checkmark       &     &  42.45  &  13.44\\
 & \checkmark & \checkmark &  43.07       & 13.93 \\
\checkmark &      &     & 41.47      & 13.45  \\
\checkmark & \checkmark &     & 43.10   & 13.68    \\ 
\checkmark & \checkmark  & \checkmark & \textbf{44.23}        & \textbf{14.81}  \\
\bottomrule[.05cm]
\end{tabular}
\caption{\textbf{Ablation study on each module in LOMA.}}
\label{tab:main_module}
\end{table}

\subsection{Comparisons with the State-of-the-Art Methods}
\begin{figure*}[t]
\centering
\includegraphics[width=0.85\linewidth]{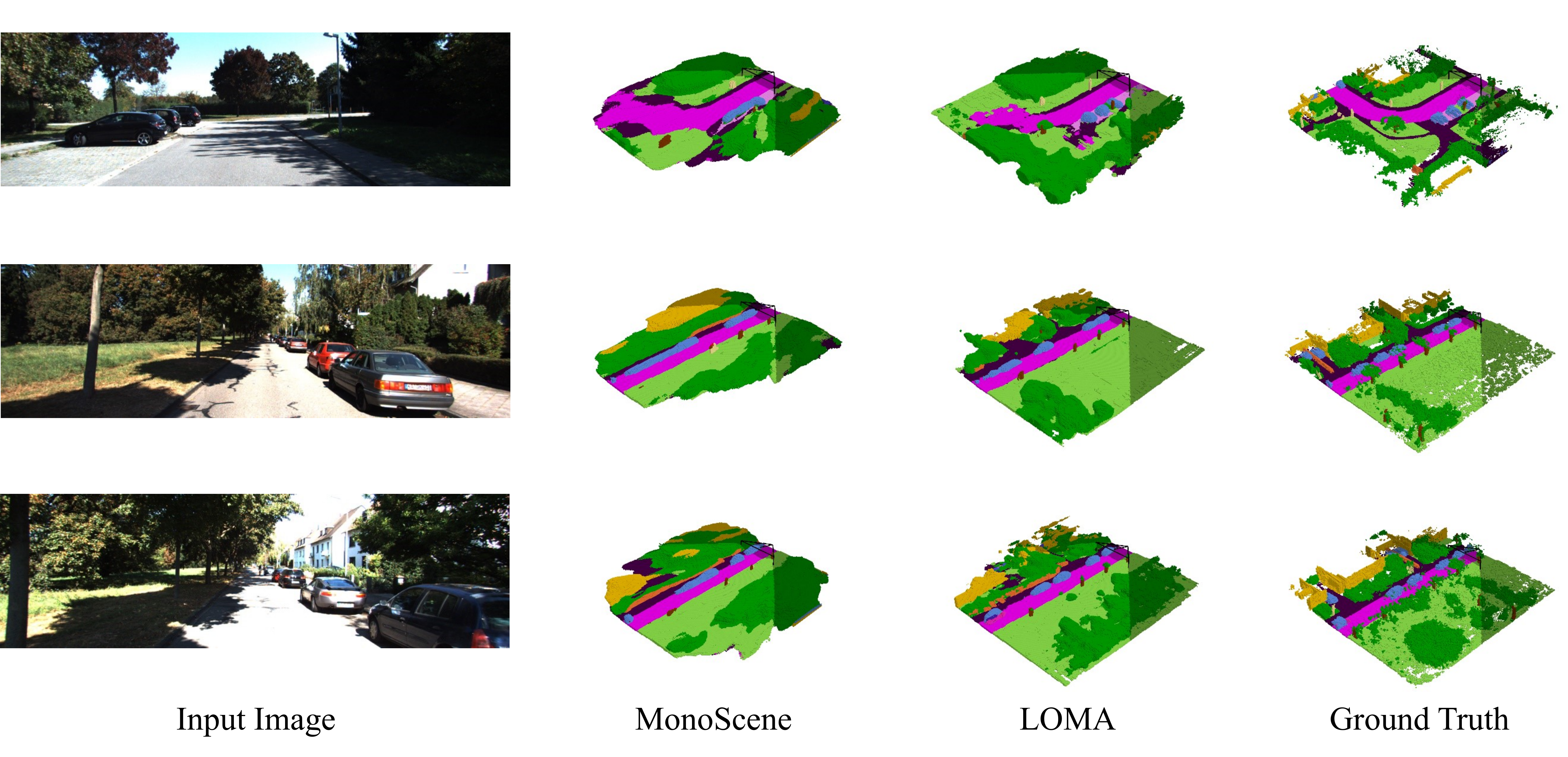}
\caption{\textbf{Qualitative visualizations on SemanticKITTI val}. Our proposed LOMA generates more refined predictions for objects and also preserves organized designs for structures.}
\label{fig:val_vis}
\end{figure*}

As shown in Table.~\ref{tab:sem_kitti_test} and Table.~\ref{tab:kitti_360_test}, our method shows superior performance on both SemanticKITTI and SSCBench-KITTI360 datasets. 
For the SemanticKITTI dataset, LOMA achieves the best mIoU and the second IoU performance. 
In the semantic comparison, our proposed LOMA shows superior prediction in several common classes, such as car, road, vegetation, and building. 
This phenomenon can be attributed to the fact that language contains rich semantic cues about these classes that improve prediction accuracy.
Meanwhile, we also notice that compared to the previous state-of-the-art method Symphonies, our approach not only improves semantic prediction (+0.06 points) but also enhances occupancy prediction (+0.82 points).
We believe that the results demonstrate our hypothesis that language not only directly provides rich semantic priors, but also implicitly guides geometric perception. Moreover, we also visualize our method in SemanticKITTI val, as shown in Figure.~\ref{fig:val_vis}. 

In the SSCBench-KITTI360 benchmark, compared to Symphonies, we improve the occupancy prediction by 2.23 points, proving that language could provide the geometric prior. We also achieve 18.28 mIoU in semantic prediction. 

\subsection{Ablation Studies}
To further analyze the effectiveness of each module in the proposed framework, we conduct ablation studies on the SemanticKITTI validation set.

\begin{table}[t]
\centering
\renewcommand{\tabcolsep}{1.2pt}
\renewcommand\arraystretch{0.8}
\begin{tabular}{cc|cc}
\toprule[.05cm]
VL-Vision Feature.      & VL-Language Feature.   &  IoU   & mIoU\\
\midrule
     &      &43.07    &13.93 \\
 \checkmark    &      &43.75    &14.14 \\
      & \checkmark    & 43.92  & 14.34  \\
 \checkmark  & \checkmark & \textbf{44.23}        & \textbf{14.81}  \\
\bottomrule[.05cm]
\end{tabular}
\caption{\textbf{Ablation study on VSG.}}
\label{tab:vsg}
\end{table}

\subsubsection{Ablation on Main Modules.}
Table.~\ref{tab:main_module} presents the breakdown analysis of various architectural components within LOMA. Without any modules, the baseline model achieves 40.42/13.36 in IoU/mIoU. With the proposed TFM, we improve the IoU and mIoU by 2.03 points and 0.08 points respectively.
This improvement indicates that our proposed TFM can effectively perform 3D-to-3D feature propagation from non-empty voxels to empty voxels. However, due to the lack of sufficient semantic perception ability, relying solely on vision features for semantic prediction is still limited.
By utilizing the high-level semantic information in a multi-scale manner, we further improve the IoU and mIoU by 0.62 points and 0.49 points.
Additionally, we add the language information to the baseline model by the proposed VSG module, leading to improvements of 1.05 and 0.09 points. 
By adding TFM and MS-TFM, we ultimately achieve the best performance with 44.23/14.81 in IoU/mIoU. Compared to the model without language, we improve the performance by 1.16 IoU and 0.88 mIoU, demonstrating the effectiveness of the language prior.

\subsubsection{Ablation on the VSG.}
We conduct an ablation study of VSG in Table.~\ref{tab:vsg} to verify the specific effects of VLM. 
We remove the language feature and vision feature from VLM, leading to 43.75/14.14 and 43.92/14.34 respectively. The two results are both better than the model without the VSG module. 
We believe that the large VLM could output better semantic vision features to help the prediction.
Meanwhile, compared to VL-vision, the VL-language achieves better performance in both semantic prediction and occupancy prediction, showing that language plays a more important role. The result also aligns with our motivation, that is, the language features in the large VLM can assist us in 3D semantic occupancy prediction.

\begin{table}[t]
\centering
\renewcommand\arraystretch{0.7}
{
\begin{tabular}{l|ccrr}
\toprule[.05cm]
Method  &IoU  & mIoU &Param & FLOPs \\
\midrule
Conv               & 40.98    & 10.89& 17.8M&34.0G     \\
Deformable  &  43.70  & 14.51 & 17.5M  &26.8G  \\
Swin2D  & 43.21   & 13.70  & 18.6M  &33.7G   \\
Swin3D &   42.57   &13.73  &0.5M  & 69.1G  \\
\hline
Ours   & \textbf{44.23}  & \textbf{14.81}   &  17.4M  & 26.9G   \\
\bottomrule[.05cm]
\end{tabular}
}
\caption{\textbf{Ablation study on components in TFM.}}
\label{tab:tmf}
\end{table}

\subsubsection{Ablation on the TFM.}
To explore the 3D-to-3D feature propagation, we compared different feature operations in the proposed TFM. We replace the SSM block in our tri-plane design with other operations, such as conv, deformable attention and swin transformer. Meanwhile, we also directly perform 3D feature interaction with swin3D transformer.
As depicted in Table.~\ref{tab:tmf}, compared to these competitors, our SSM-based method not only achieves the best performance but also incurs limited FLOPs. 
The result validates the efficiency of global modeling in our method.
Meanwhile, the main source of parameters lies in the 3D-to-2D and 2D-to-3D linear layers, but not the feature processing module. 

\begin{table}[t]
\centering
\renewcommand\arraystretch{0.7}
{
\begin{tabular}{l|ccrr}
\toprule[.05cm]
Method  &IoU  & mIoU &Param &FLOPs \\
\midrule
Dense               & 42.97     &14.70 &4.4M & 72.5G    \\
Ours               & \textbf{44.23}     &\textbf{14.81}& 4.4M & 42.4G    \\ \hline
$[1]$  & 43.10  & 13.68  & 17.6M &43.8G    \\
$[1, 1/2]$ & \textbf{44.23}     &\textbf{14.81}& 57.2M& 219.9G    \\
$[1, 1/2, 1/4]$   & 42.82  & 13.88   & 146.2M   & 301.7G    \\
\bottomrule[.05cm]
\end{tabular}
}
\caption{\textbf{Ablation study on architectural components in Multi-scale Designs.}}
\label{tab:ms_tmf}
\end{table}

\subsubsection{Ablation on the MS-TFM.}
Table.~\ref{tab:ms_tmf} compares the designs in our multi-scale extension. We first replace the sparse conv with regular dense conv, leading to a decrease of 1.26/0.11 points and a higher computation cost. Compared to dense conv, sparse conv can more effectively model the occupied geometry by extracting vital signs from non-empty voxels. We further compare different scales in a multi-scale manner. Compared to single scale or three scales, our two-scale setting performs better performance in both semantic prediction and occupancy prediction. We analyze that a single scale may lack some high-level semantic details, while too many scales may lead to an abundance of noise interference.

\section{Conclusion}
In this paper, we introduce a new vision-language framework for the vision-based 3D semantic occupancy prediction, named LOMA. By integrating the language information from the proposed VL-aware Scene Generator module, LOMA gets better geometric and semantic perception. Meanwhile, to have a global feature modeling with limited computation cost, we also introduce the Tri-plane Fusion Mamba block to perform the 3D-to-3D feature propagation. By extending the TFM module into the multi-scale approach, our method performs better prediction. In future work, we would like to equip LOMA with more modalities to achieve better and more accurate predictions. 

\appendix

\section{Acknowledgments}
This work was supported in part by the National Natural Science Foundation of China under Grants 62073066, in part by the Fundamental Research Funds for the Central Universities under Grant N2226001, and in part by 111 Project under Grant B16009.




\bibliography{aaai25}

\end{document}